\documentclass[10pt,twocolumn,letterpaper]{article}

\usepackage{amsmath,amssymb}
\usepackage{graphicx}
\usepackage{verbatim}
\usepackage{color}
\usepackage{algorithm, algorithmic}
\usepackage{subfig}
\usepackage{hyperref}
\usepackage{xcolor}

\newcommand{\nb}[3]{{\colorbox{#2}{\bfseries\sffamily\scriptsize\textcolor{white}{#1}}}
{\textcolor{#2}{\sf\small\textit{#3}}}}
\newcommand{\pp}[1]{\nb{Pavlos}{red}{#1}}

\begin{document}

\title{T-CGAN: Conditional Generative Adversarial Network for Data Augmentation in Noisy Time Series with Irregular Sampling}

\author{Giorgia Ramponi
\thanks{Politecnico Milano, DEIB. Milano, Italy.} \and
Pavlos Protopapas \thanks{Harvard University, Cambridge, MA, USA.} \and
Marco Brambilla  \thanks{Politecnico Milano.  Milano, Italy.} \and 
Ryan Janssen \thanks{Harvard University,  Cambridge, MA, USA.}}

\date{}

\maketitle







\begin{abstract} \small\baselineskip=9pt

In this paper we propose a data augmentation method for time series with irregular sampling, Time-Conditional Generative Adversarial Network (T-CGAN). Our approach is based on Conditional Generative Adversarial Networks (CGAN), where the generative step is implemented by a deconvolutional NN and the discriminative step by a convolutional NN. Both the generator and the discriminator are conditioned on the sampling timestamps, to learn the hidden relationship between data and timestamps, and consequently to generate new time series. 

We evaluate our model with synthetic and real-world datasets. For the synthetic data, we compare the performance of a classifier trained with T-CGAN-generated data, against the performance of the same classifier trained on the original data. 
Results show that classifiers trained on T-CGAN-generated data perform the same as classifiers trained on real data, even with very short time series and small training sets.  
For the real world datasets, we compare our method with other techniques of data augmentation for time series, such as time slicing and time warping, over a classification problem with unbalanced datasets. 
Results show that our method always outperforms the other approaches, both in case of regularly sampled and irregularly sampled time series. We achieve particularly good performance in case with a small training set and short, noisy, irregularly-sampled time series.

\end{abstract}
\section{Introduction.}

Time series data are ubiquitous. Some examples include stock prices \cite{KAZEM2013947}, currency exchange rates, sales data \cite{CHEN20107696}, biomedical measurements \cite{Wacker2016TimefrequencyTI}, astronomical data \cite{Rebbapragada2009}and weather data collected over time \cite{CHEN20112856}.


However, for many applications \textbf{only small labeled datasets are available} for training machine learning methods and this often results in low performance of the task in hand. 

To solve the small amount of data available in the training set, the simplest solution is to 
collect more labelled data, but often this task is unfeasible or too expensive.
Another solution is to perform data augmentation. Data augmentation is the technique of creating synthetic data to augment the size of a dataset. 

In the case of classification, training on augmented datasets can increase the performance of the classifier.
In fact, it is well-known that too small of training dataset can cause overfitting \cite{10.1007/978-3-319-08010-9_33} and that the overfitting decreases with the increase of the size of the dataset \cite{Nonnemaker} \cite{Brain}.

While many dataset augmentation techniques exist for image data (for instance, images can be flipped, translated or rotated \cite{Wang2017}), these methods do not generalize well to time series. For time seres data a simple visual comparison cannot confirm if the transformation change the nature of the time series, instead it could be done easily for an image.
This is the main reason why data augmentation for time series classification has been limited to mainly two relatively simple techniques: time slicing and time warping \cite{LeGuennec}.

Time slicing is a method inspired by computer vision communities, which consists in cropping slices from time series and performing classification at the slice level. This method has been introduced for time series in \cite{Cui}. 
Time slicing can be less effective, because cutting the time series tends to remove temporal correlation in the data.

Time warping is a time-series specific method consisting of {\em warping} a randomly selected slice of a time series by stretching it, i.e., speeding it up or slowing it down \cite{LeGuennec}.
This method, in theory, should not alter the distribution of the data significantly. Its main problem is that it does not generalize well, and in some cases (such as astronomy), the time scale has significant physical meaning.  As a result, the time warped data may have a very different interpretation. 


These two methods however work properly only over regularly sampled time series. In real settings however, irregular time sampling is a critical problem in data analysis. Irregular sampling can occur because of several issues, such as scheduling patterns, technical faults in sensing devices, imprecision of the sensors and timing devices, or human errors.

The result is a time series where the data points' position in time is irregular. 
The irregular time sampling can result in substantial uncertainty about the values of the underlying time series, thus making it more difficult to mimic the time series in a realistic manner. Furthermore, irregularity makes the data difficult to deal with using standard classification methods that assume fixed-dimensional feature spaces, also because it prevents the application of basic data augmentation approaches. 

\textbf{In this paper we propose \emph{Time-Conditional Generative Adversarial Network (T-CGAN)}, a method aiming at generating new irregularly-sampled time series, with the objective of augmenting unbalanced data sets in time series classification problems.} 

Given a dataset $S$ of irregularly sampled time series, our goal is to generate new time series which mimic the ones in the dataset $S$ in a realistic way. Note we neither generate missing data points in the time series, nor regularize irregular time stamps. We generate new irregularly-sampled time series instead.
 To  obtain  this, we implement a time-aware conditional generative adversarial network (T-CGAN). Our  method  works  by  conditioning  the  generator  and  the discriminator with the timestamps. The goal of T-CGAN is to discover the latent space of the time series in order to mimic the time series dynamics.
 We aim at covering a realistic problem setting, and therefore we assume that the time series are noisy.
 
 We  evaluate  our  model  in  two  different  scenarios:  with synthetic datasets and with three real world datasets with unbalanced classes.  
 
 In the synthetic scenario we compare the performance of a classifier trained with data generated by T-CGAN against the performance of  the  same  classifier  trained  on  the  original data. The classifier is implemented with a simple convolutional neural network (CNN) and the test is always run on the original data. Results show that T-CGAN based training enables good results of the classifier, even with very short time series and small training sets.
 
 In the real world  experiment, we consider an unbalanced-class classification problem and we use the T-CGAN to generate time series in the class which features the smaller training set, so as to move to a perfectly balanced setting. 
 Over the real dataset classification problems,  we also  compare  our  method  with  state-of-the-art techniques of data augmentation for time series, such as time slicing  and  time  warping. 

Results show that our method always performs better than the other approaches, both in case of regularly sampled and irregularly sampled time series. We achieve  particularly good performance in case of small training set and short noisy irregularly-sampled time series.

The paper is organized as follows: Section \ref{sec:relwork} discusses the related work, Section \ref{sec:bkg} presents the background, Section \ref{sec:model} describes the T-CGAN model; Section \ref{sec:exp} reports on our experiments; and Section \ref{sec:concl} concludes.

\section{Related work}
\label{sec:relwork}
Time series generation is a specific application problem of the broad field of sequential data generation, where a sequence is dictated by a temporal variable. Sequence generation may be applied to continuous or discrete elements. 
In this section, we discuss the related work for data augmentation techniques for sequential data, and in particular data augmentation for time series.

\subsection{Discrete Sequential Token Generation}

The major interest in sequential data generation has been on discrete useful tokens in fields like NLP, where the challenge is to generate appropriate sequences of words. For example, Yu et Al. \cite{Yu2016} proposed a GAN-based approach for natural language processing to generate sequences of discrete tokens using a GAN trained by a reinforcement learning approach. Recently, conditional GAN architectures have also been used in NLP, including translation \cite{DBLP:journals/corr/YangCWX17} and dialogue generation conditioned on a particular sentiment \cite{li2017adversarial}. These methods aim to infer the next value of the sequential series, but they don't prove the capacity on generating new data in order to augment the dataset.

\subsection{Temporal Data Generation using Generative Adversarial Networks}
In 2018 Hyland et al. \cite{20.500.11850/236194} proposed a GAN based on a Recurrent Neural Network both for the generator and the discriminator, in order to produce realistic real-valued multi-dimensional time series for medical data. In this work they introduced the \textit{train on synthetic, test on real} methodology to test the quality of the generated data. 

Mogren et al \cite{Mogren2016} proposed a solution to generate continuous-valued sequences that aims to produce polyphonic music using a GAN with LSTM generator and discriminator.  
In this work they succeeded in producing data which are realistic, but they did not consider cases with irregular time sampling and noisy signal. 

More recently, other works  aimed to generate new data from data sources with missing observations. In the recent work of Yoon et al.\cite{pmlr-v80-yoon18a}, they proposed a model to reconstruct missing data where the generator (G) received as input the real data vector, imputes the missing components conditioned on what is actually observed, and outputs a completed vector. The main differences between the model proposed by Yoon et Al. and our approach are two: the aim of this research is not to create new data from data with noise but to reconstruct missing points; their approach is not time series dedicated.

\subsection{Data Augmentation}
Data augmentation is a data generation strategy typically used for supervised problems in  machine learning, with the objective of producing relevant data points for improving the learning of ML solutions (for instance, classifiers).
The most commonly used data augmentation method for time series is the time slicing window technique, originally introduced for deep CNNs  in \cite{Cui2016MultiScaleCN}. This method takes inspiration from computer vision \cite{Zhang2016} and when used for images it can guarantee, at some levels of cropping, that an image divided in slices maintains the same information as the original image. The method does not give the same guarantees for time series data, because it is not obvious that the discriminator information is maintained when a region of the time series is cropped. Nevertheless, this method was used in several time series classification problems,  such as in \cite{KVAMME2018207}, where they used CNNs  to improve mortgage delinquency prediction with  using customers historical transactional data, and in \cite{Lines2015} where it was used to improve the accuracy of a Support Vector Machines classifier for electroencephalographic time series data. 

It is important to note that with the time slicing technique, the model classifies each sub-sequence alone and then finally classifies the whole time series using a majority voting approach over the set of sub-sequences.  This can cause the loss of important information about the time series data distribution. Contrarily, the method that we propose in this paper does not crop time series into shorter subsequences, using the discriminator properties from the whole time series.

Other techniques for augmenting time series data have been proposed in literature, such as jittering, scaling, warping and permutation. 
For example the authors in \cite{Um} created an innovative data augmentation method for wearable sensor time series data to classify the motor state of Parkinson's disease patients.
In \cite{LeGuennec}, the authors propose a method for data augmentation that is a mixture between time slicing and time warping, using the time warping technique to create new data and time slicing to create time series of the same length. This method was used to improve the classification of their deep CNN for time series classification. 
Recently Fawaz et al.\cite{fawaz} used dynamic time warping to augment time series dataset in order to increase the classification performance of a deep residual network. This work shows how data augmentation can drastically improve the classification accuracy.

\section{Technical Background}
\label{sec:bkg}
In this section we introduce the technical background on Conditional Generative Adversarial Networks (GANs) that we use in the rest of the paper.

GANs were introduced by I. Goodfellow \cite{NIPS2014_5423} as a model to train a generative model. GAN model consists of two models which play a two-player min-max game:
\begin{itemize}
    \item a generative model, G, that has the goal of capturing the data distribution;
    \item a discriminative model, D, that has the task of identifying if a sample comes from the training data or from G.
\end{itemize}
The generative model G has to learn a distribution $p_g$ 
over data $x$, by building a mapping function from a prior noise distribution $p_z(z)$ to the data space, $G(z;\theta_g)$, where $\theta_g$ are the parameters of the model, e.g. the multilayer perceptrons weights implementing $G$. 

The discriminator $D(x;\theta_d)$ instead is a second multi-layer perceptron implementing a binary classifier, which outputs a single scalar representing the probability that $x$ came form training data rather than $p_g$.

The two models are trained together to play the following two-player min-max game:

\begin{multline}
    \min_{G} \max_{D} V(D,G) = \\
    \mathbb{E}_{x \sim p_{data}(x)}[\log D(x)] +
    \mathbb{E}_{z \sim {p_z}(z)}[\log (1-D(G(z)))]
\end{multline}

This model can be extended to a conditional model \cite{DBLP:journals/corr/MirzaO14} if both the generator and the discriminator are conditioned on some extra information $y$.
The conditioning is performed by feeding $y$ into both the discriminator and the generator as additional input:
\begin{itemize}
    \item in the generator, the prior input noise $p_z(z)$ and $y$ are combined in a joint hidden representation, and the adversarial training framework allows for considerable flexibility in how this hidden representation is composed;
    \item in the discriminator, $x$ and $y$ are presented as inputs and to a discriminative function.
\end{itemize}
The objective function of the two-player minimax game is now:
\begin{multline}
   \min_{G} \max_{D} V(D,G) =\\
    \mathbb{E}_{x \sim p_{data}(x)}[\log D(x|y)] +  \mathbb{E}_{z \sim p_z(z)}[\log (1-D(G(z|y)))]
\end{multline}

\section{T-CGAN Model}

\begin{figure}
    \centering
    \includegraphics[width=\columnwidth]{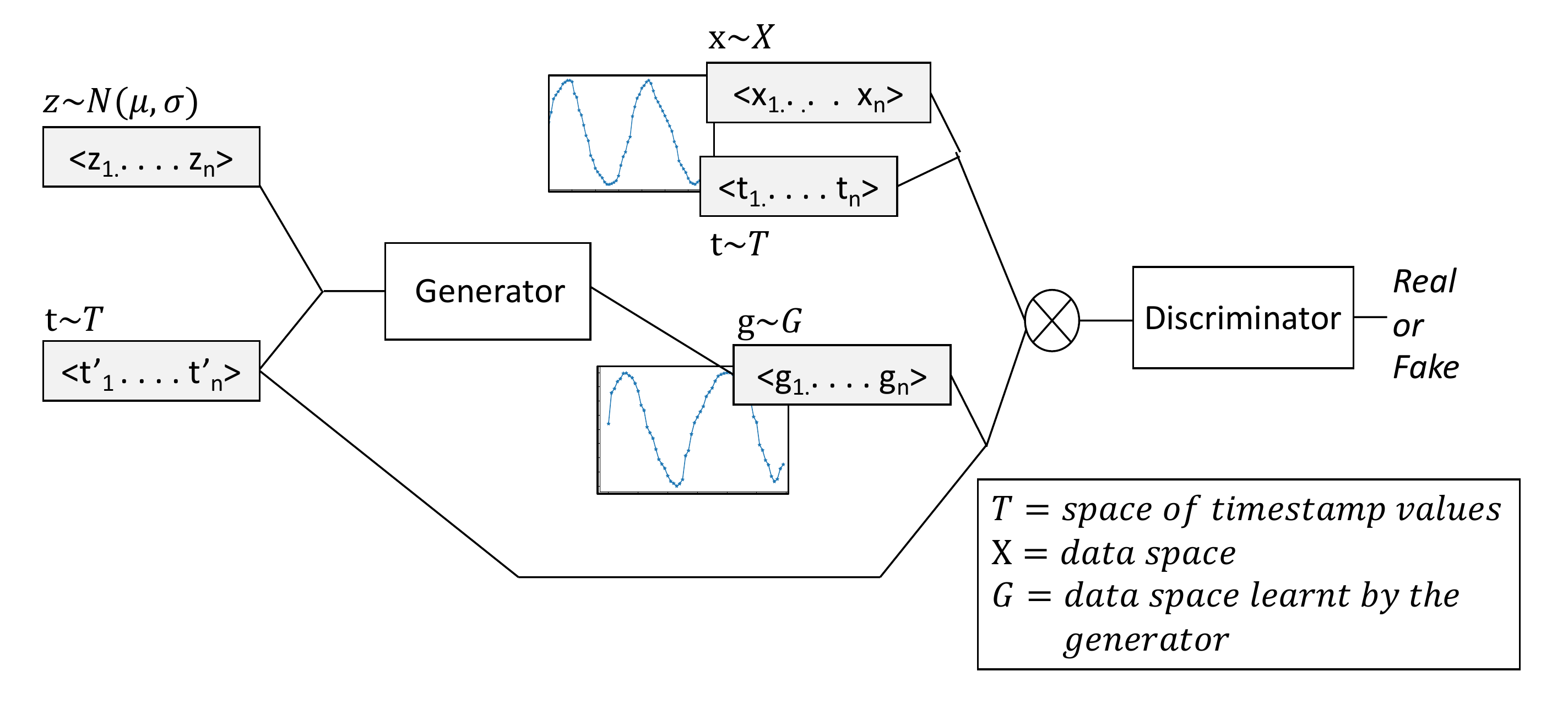}
    \caption{Model of Time-Conditional Generative Adversarial Network (T-CGAN).} 

    \label{model}
\end{figure}
\label{sec:model}

In this section, we propose a model based on CGAN framework to solve the problem addressed in the introduction. The model that we are proposing is a Conditional Generative Adversarial Network Model (see Figure \ref{model}) composed by two CNNs, one for the generative part (G) and one for the discriminative part (D).

We define now the input and output space of our model, each with an associated probability distribution:
\begin{itemize}
    \item $Z$ is a noise space used to seed the generative model.
    Samples of $z \in Z$ are sampled from a noise distribution $p_z(z)$. In our experiments $p_z$ is a simple Gaussian noise distribution with mean equals to $0$ and standard deviation equals to $1$.
    \item $T$ is the space of timestamps used to condition the generative and the discriminative model. 
   
    \item $X$ is the data space which represents a time series output from the generator or input to the discriminator. Values are the data of the time series. Using the time series in the training data and their associated conditional 
    data, we can define a density model $p_{data}(x,t)$. This is exactly the density model we wish to replicate with the overall model in this paper. 
\end{itemize}
We can define in this way the objective function of our model as:
\begin{multline}
   \min_{G} \max_{D} V(D,G) =\\
    \mathbb{E}_{x \sim p_{data}(x)}[\log D(x|t)] +  \mathbb{E}_{z \sim p_z(z)}[\log (1-D(G(z|t)))]
\end{multline}

where $t=<t_1,...,t_n>$ is a sorted vector of timestamps sampled at random from $T$. Notice that the model is also able to generate new time series corresponding to timestamps not present in the training set.

\subsection{Generative Network}
We can define the generative network as a function $G: (Z,T)\rightarrow X$, which has as input data $z\in Z$ and $t \in T$ and outputs a time series $x\in X$. The generative network G is a CNN, which takes as input noise and timestamps and outputs the value of the time series for the given timestamp.
This transformation  is done through four convolutional transpose (or deconvolution) layers with ReLUs activation functions and batch normalization at each layer except for the last one. 

The generative network adjusts its parameter to minimize $\log(1-D(G(z|t)))$, where $z \in Z$ is the noise vector and $t \in T$ is the \textit{timestamp} vector.

\subsection{Discriminative Network}
The Discriminative network implements a function: \\ $D: (X,T)\rightarrow [0,1]$. This network takes as input real data $x\in X$ or generated data $g \in G$, and their associated timestamps $t \in T$, and gives as output a binary value, deciding whether the data is real or generated. It is composed by two layers of convolution, each followed by a max-pooling layer. At the end there is a fully connected layer.

The Discriminative Network adjusts its parameter to maximize $\log(D(x|t))$, where $x$ is the time series vector and t is the \textit{timestamp} vector.

\section{Experiments}
\label{sec:exp}
In this section we validate the performance of T-CGAN using synthetic data and three real-world datasets. 
We first describe the datasets used for the experiments \cite{bagnall16bakeoff}, in Section  \ref{sec:datasets}.

In Section \ref{sec:qualitative}, we report on the results of the experiment over synthetic irregularly sampled data, by comparing the performance of a binary classifier trained over the original and the generated data to distinguish two curve types (sine and sawtooth).

In Section \ref{sec:regular}, we quantitatively evaluate the performance of T-CGAN in improving the classification using three real-world datasets of regularly sampled data
\cite{bagnall16bakeoff} modified to create unbalanced datasets. We compare the results of data augmentation performed with T-CGAN against the   the ones obtained with time slicing and time warping methods for data augmentation.

 In Section \ref{sec:irregular} we evaluate the performance of T-CGAN against time slicing and time warping, with the same setting as before but considering the case of irregularly sampled datasets. In this setting, the datasets are created starting from the original three real datasets used above, by randomly removing different shares (from 10\% to 40\%) of points from each time series.
 We then applying T-CGAN to generate new time series.

We run all the experiments with 10-fold randomization and we report the Area Under the Receiver Operating Characteristic Curve (AUROC) as the performance metric (mean value across the repeated experiments, along with the standard deviation).

\subsection{Datasets}
\label{sec:datasets}
This section describes the synthetic and real world datasets used for experiments.

\begin{figure}
    \centering
    \includegraphics[width=\columnwidth]{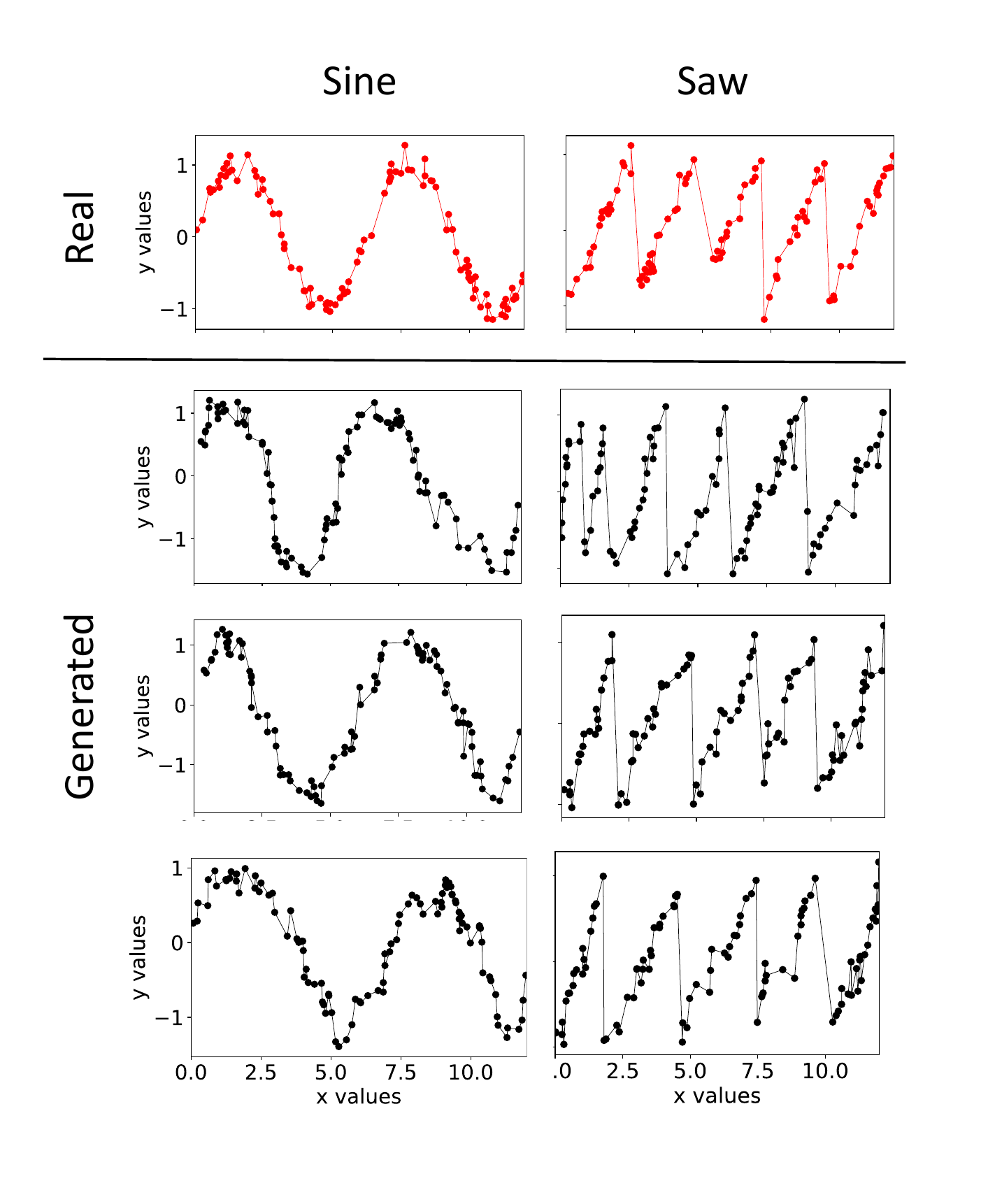}
    \caption{Example of real curves (noisy and irregularly sampled) and generated curves for sine and sawtooth signals. }
    \label{fig:Ngen-curves}
\end{figure}

\subsubsection{Synthetic data}

We construct as realistic input data one synthetic datasets, composed by two classes: sine waves and sawtooth waves (see Fig.~\ref{fig:Ngen-curves}). 
The parameters considered for producing the synthetic input data for  the experiment are:
    the \textit{training set size} ($S$), i.e., the number of curves used to train the T-CGAN; and the
    \textit{time series length} ($L$), i.e., the length of each time series.

Each sine and saw wave is constructed as follows:
\begin{itemize}
    \item We select uniformly at random $L$ points between $[0,12]$, which are the $t = <t_1,...,t_{L}>$ timestamps;
    \item 
    We define an amplitude $a = 1 + d_a$ , where $d_a$ is selected at random from a normal distribution with mean $ \mu = 0$ and standard deviation $ \sigma = 0.1$;
    \item We assume period $ P = 2\pi$, with phase shift $= 0$ and vertical shift $= 0$;
    \item We add a Gaussian noise (with $\mu = 0$ and $\sigma = 0.1$) to each point of the series.
\end{itemize}
We repeat the process $S$ times for generating all the needed curves. 
Figure \ref{fig:Ngen-curves} reports  exemplary real and T-CGAN-generated curves for the sine and sawtooth classes.

\subsubsection{Real world data}
\label{real_datasets}

The real world datasets are taken from Time Series Classification UCF \cite{bagnall16bakeoff}.  Table \ref{tab:dataset}  summarizes the characteristics of the three datasets used for the experiment. Note that we use shorter time series than the original ones, and we create artificially irregular sampling by removing points at random from the original series. 

\begin{table}[t]
    \caption{Characteristics of the three datasets used to evaluate our method. 
    }
    \label{tab:dataset}
\
    \centering
    \footnotesize
    \begin{tabular}{l|ccc}
    & \multicolumn{3}{c}{DATA SETS}\\
         \textbf{} & \textbf{Starlight} & \textbf{Power} & \textbf{ECG} \\
         \textbf{} & \textbf{Curves} & \textbf{Demand} & \textbf{200} \\
         \hline \hline
        
         Timeseries length  &80& 24 & 96  \\
          \hline
        
         1st training class size &20 & 100 & 27\\
        
         2nd training class size &200& 200 &93 \\
          \hline
        
         {Test set class size} & 100 & 200 & 40
    \end{tabular}
\end{table}

\begin{figure*}
    \centering
    \small
    \includegraphics[width=.52\columnwidth]{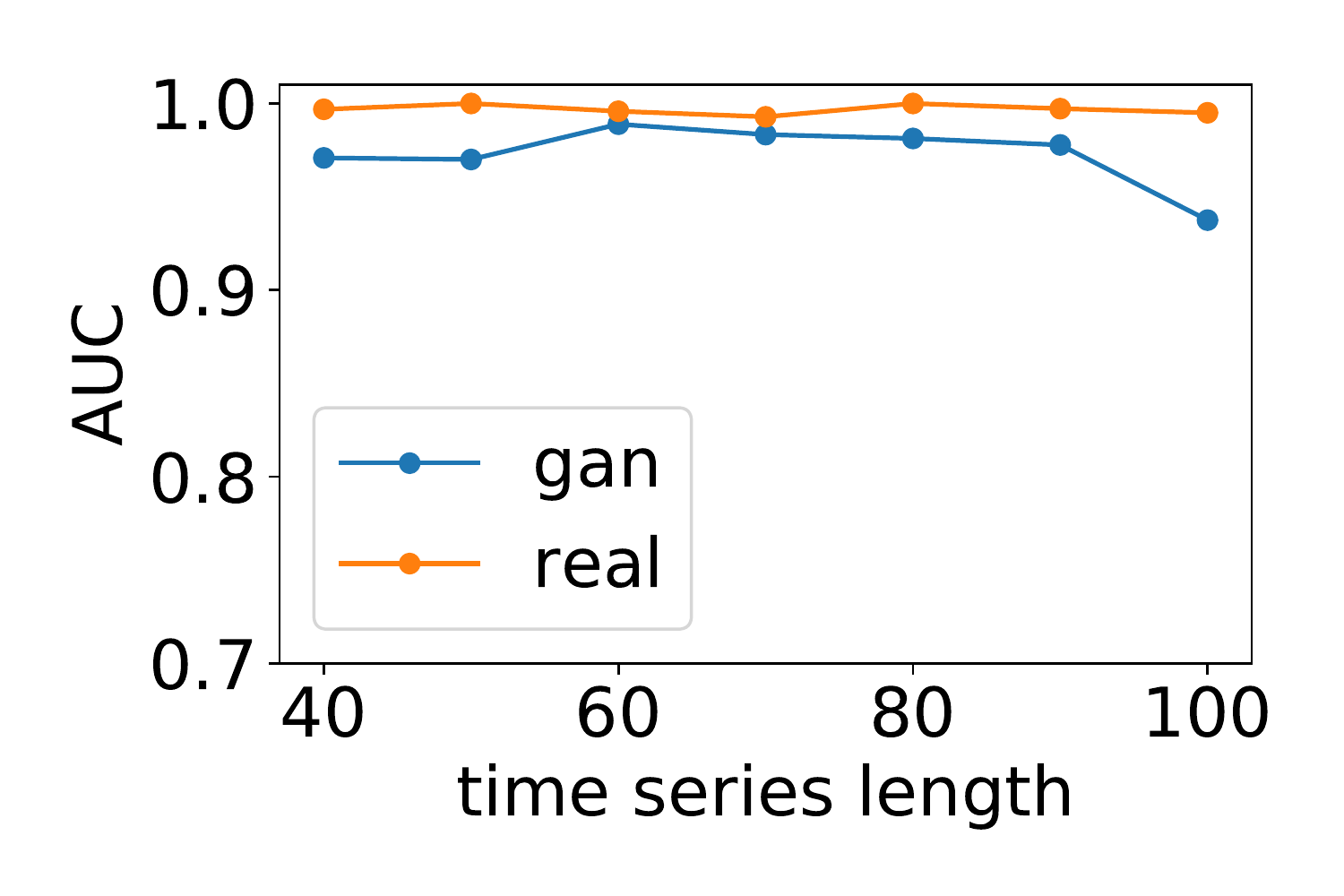}
    \includegraphics[width=.42\columnwidth]{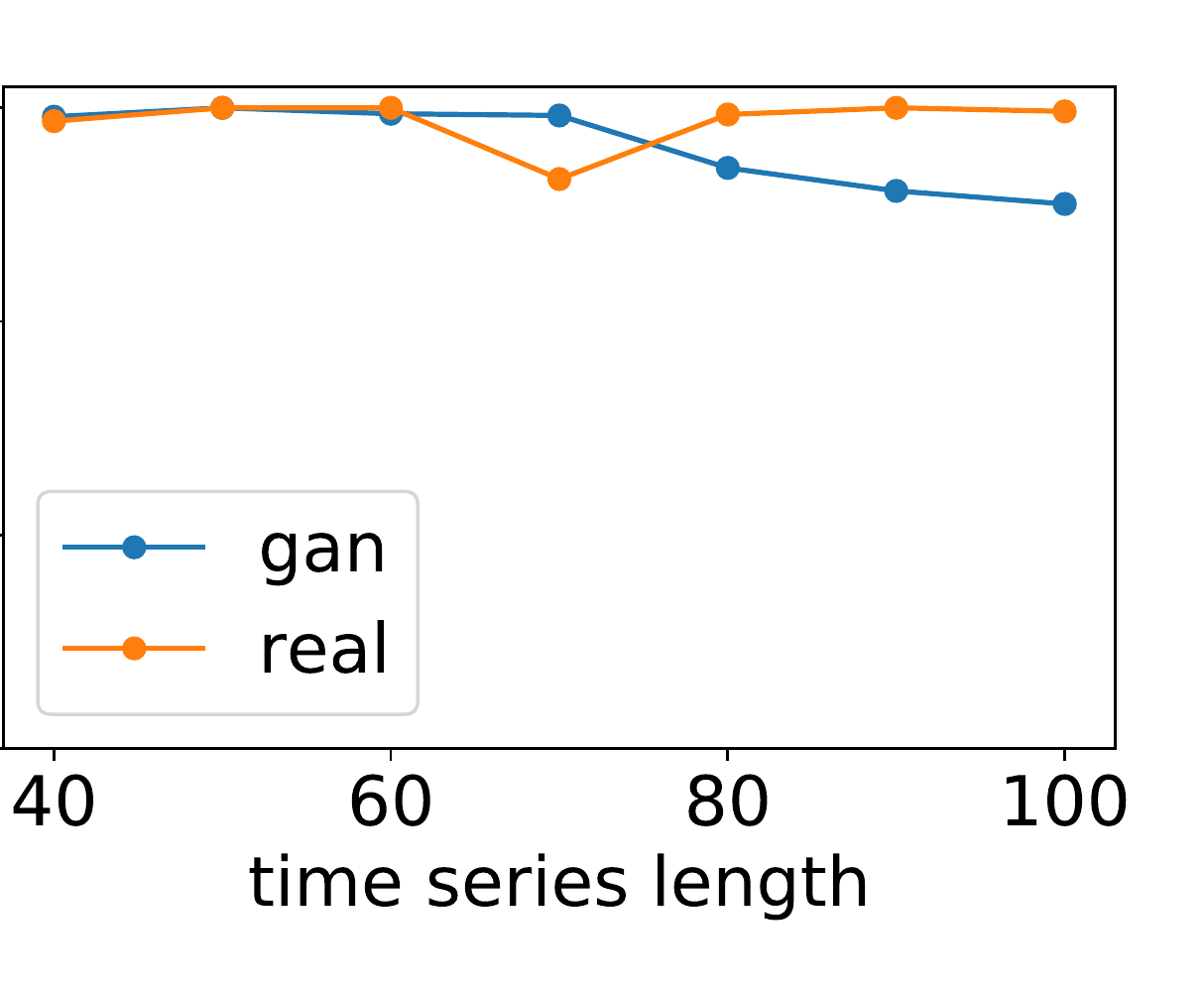}
        \includegraphics[width=.42\columnwidth]{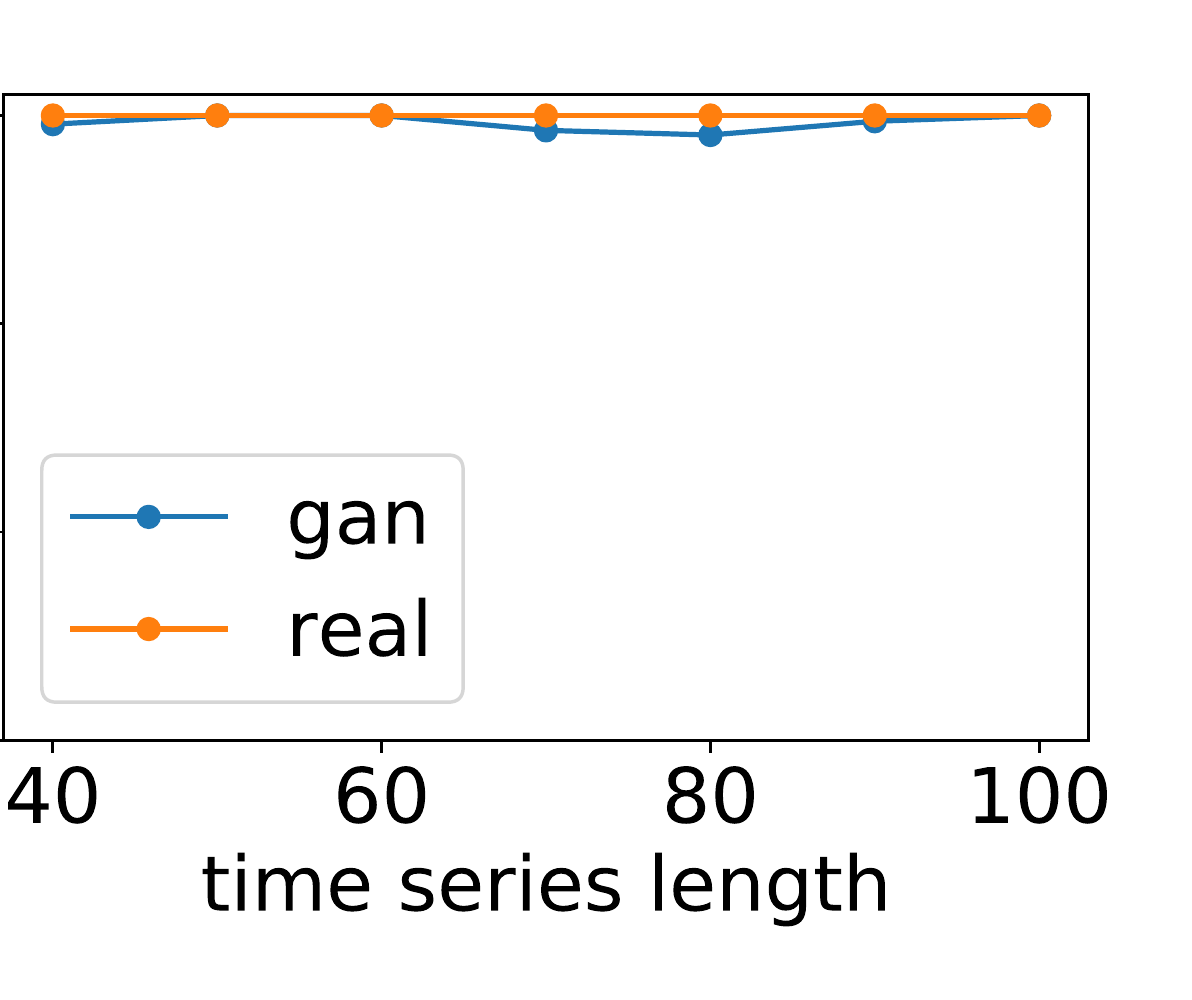}
    \includegraphics[width=.42\columnwidth]{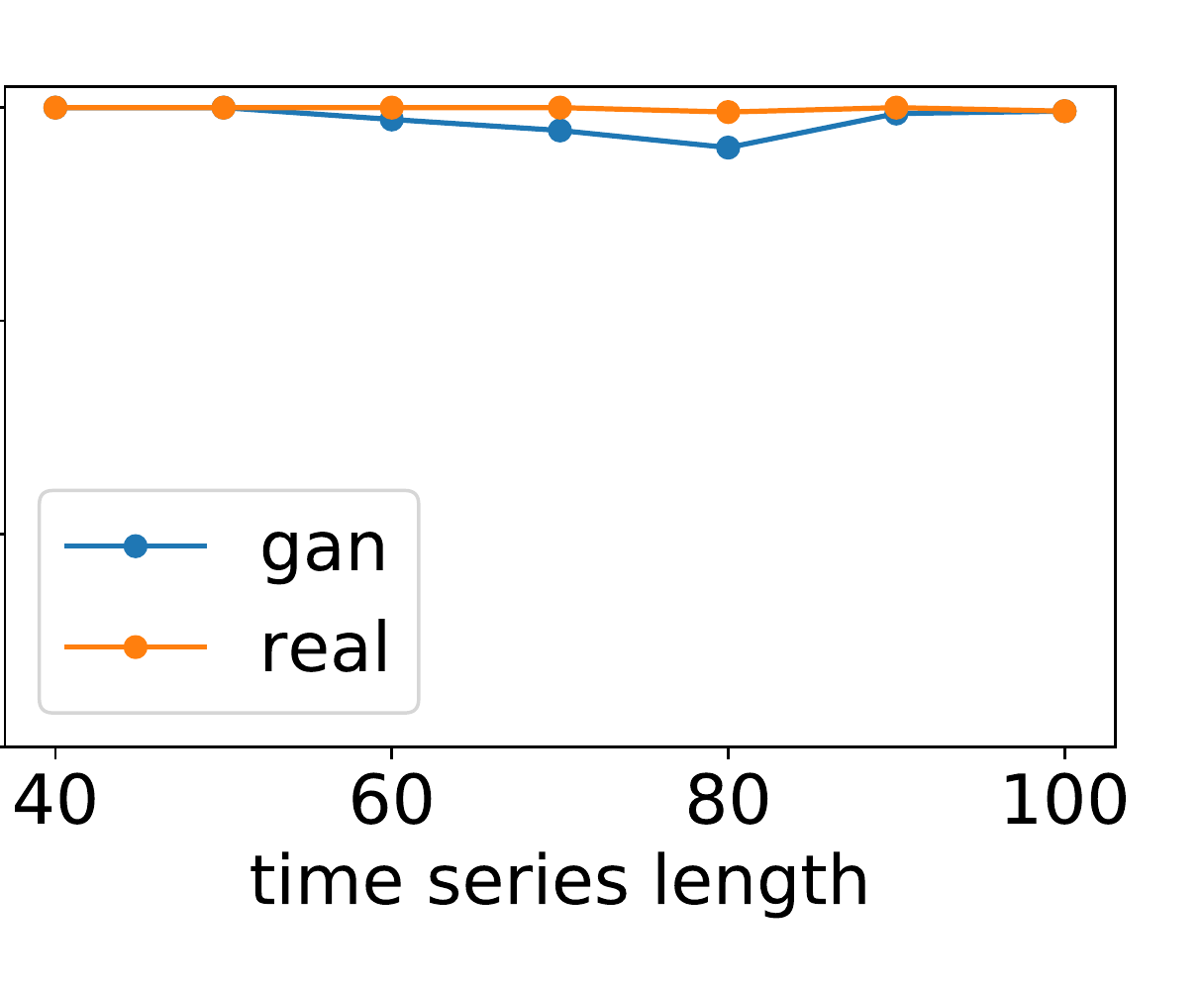}

     (a) Training set size = 40  
     \hspace{6pt} 
  (b) Training set size = 60  
  \hspace{6pt}
  (c) Training set size = 80  
     \hspace{6pt} 
  (d) Training set size = 100  
    \caption{
    AUROC for classification problem over synthetic data with different training set size (40,60,80,100 curves) and different length of synthetic data (considered length values: $40, 50, 60, 70, 80,90,100$), with the classifier trained on real data and trained on  T-CGAN generated data.}
    \label{fig:accuracy}
\end{figure*}

\textbf{\textit{Starlight curves}}:
This dataset is composed by astronomical light curves (brightness of celestial objects). The dataset has $1000$ phase-aligned 
starlight curves of length $1024$ 
\cite{Rebbapragada2009}. The curves are divided in three classes, but we use only two classes (\textit{\#2} and \textit{\#3})
in our experiments. In the experiments we use only the first $80$ points of every curve. 

To create a dataset of unbalanced classes we use a training set of $20$  curves for class \textit{\#2} and $200$ for  class \textit{\#3}. The test set consists of $100$ curves for each class.

\textbf{\textit{Power Demand}}: The data was derived from twelve monthly electrical power demand time series in Italy \cite{Lonardi}. The classification task is to distinguish days from October to March (inclusive) and from April to September. The length of the time series is 24 data points. To create unbalanced classes we use a training set of $100$  time series from the first class and $200$ from the second class. The test set consists of $100$  curves for each class.

\textbf{\textit{ECG200}}: This dataset includes time series tracing the electrical activity recorded during one heartbeat in human hearts, with 96 data points each \cite{Olszewski:2001:GFE:935627}. The series are classified in two classes, i.e., normal heartbeat and myocardial infarction. 
To create unbalanced classes we pick a training set of $27$ time series from the first class and $93$ from the second class. The test set includes $40$ series for each class.

\subsection{Classification performance 
over synthetic data}
\label{sec:qualitative}
The first experiment aims at evaluating the quality of the data generated by T-CGAN over the synthetic dataset described in the previous section. 
To verify the quality of the curves generated by T-CGAN when fed with the synthetic data,  we apply the method introduced in \cite{20.500.11850/236194}. In particular, we consider a binary classifier implemented as a CNN for distinguishing curve classes (sine vs. sawtooth waves in our case). In addition, the classifier receives the information about timestamps, in order to calculate the relationship between values and timestamps. We compare the performance of the classifier trained on T-CGAN generated data (see Algorithm \ref{Alg})  versus the same classifier trained on real data, evaluating both classifiers on real data over performance measured according to AUROC. 

We expect that the performance of the classifier trained on generated data is comparable with the performance of the classifier trained on real data. 
If the T-CGAN trained classifier does not succeed in classifying real data, it means that the generated data points  are too different from the real ones, i.e. the T-CGAN does not succeed in mimicking the real distribution of the curves.

\begin{algorithm}[b]
\caption{Evaluate quality}
\label{Alg}
\begin{algorithmic}
\small
\STATE $train, test \gets \text{split }\textit{data}$
\STATE $curves_{gan}, labels_{gan} \gets $
\STATE \hspace{1.5cm}$ \text{T-CGAN(\textit{train.data},\textit{train.timestamps})}$
\STATE $\text{classifier.train}(curves_{gan}, labels_{gan})$
\STATE $prediction \gets $ 
\STATE \hspace{1.5cm} $\text{classifier.predict}(\textit{test.data},\textit{test.timestamps})$
\STATE $labels\_test \gets test.get\_labels$
\STATE $accuracy \gets evaluate(prediction, labels\_test)$
\end{algorithmic}
\end{algorithm}

We report the results on Table \ref{f1_synth} and in Figure \ref{fig:accuracy}. We used $S = [40,50,60,70]$ and $L = [40,50,60,70,80]$. Each experiment is repeated 10 times. 
The average AUROC values reported in the table
 show that the classifier trained on GAN-generated time series reaches comparable performance of the same classifier trained on real time series.

\begin{table*}[]
    \centering
    \small
    \begin{tabular}{c|ccccccc}
{\textbf{Training Set}} & \multicolumn{6}{c}{\textbf{Time Series Length}}\\
 \textbf{Size Type}    &      \textbf{40}  &             \textbf{50}  &             \textbf{60} &             \textbf{70}  &             \textbf{80}  &             \textbf{90}  &             \textbf{100}  \\
\hline

40 \hspace{0.2cm}  GAN     &  0.970$\pm$0.015 &  0.966$\pm$0.031 &  0.995$\pm$0.005 &  0.991$\pm$0.011 &  0.995$\pm$0.005 &      1.0$\pm$0.0 &      1.0$\pm$0.0 \\
50 \hspace{0.2cm} GAN      &  0.970$\pm$0.024 &  0.993$\pm$0.004 &      1.0$\pm$0.0 &  0.990$\pm$0.008 &      1.0$\pm$0.0 &  0.993$\pm$0.009 &      1.0$\pm$0.0 \\
60   \hspace{0.2cm}  GAN   &  0.988$\pm$0.010 &  0.991$\pm$0.006 &  0.997$\pm$0.003 &  0.988$\pm$0.010 &      1.0$\pm$0.0 &  0.988$\pm$0.015 &  0.994$\pm$0.003 \\
70  \hspace{0.2cm} GAN      &  0.983$\pm$0.006 &  0.949$\pm$0.055 &  0.996$\pm$0.003 &  0.992$\pm$0.007 &  0.992$\pm$0.007 &      1.0$\pm$0.0 &  0.989$\pm$0.010 \\
80 \hspace{0.2cm}  GAN     &  0.981$\pm$0.012 &  0.931$\pm$0.031 &  0.971$\pm$0.028 &  0.993$\pm$0.006 &  0.990$\pm$0.003 &  0.984$\pm$0.015 &  0.981$\pm$0.026 \\
90  \hspace{0.2cm} GAN      &    0.977$\pm$0.0 &  0.972$\pm$0.016 &  0.961$\pm$0.033 &  0.988$\pm$0.005 &  0.997$\pm$0.002 &  0.988$\pm$0.011 &  0.997$\pm$0.002 \\
100  \hspace{0.07cm} GAN   &  0.937$\pm$0.027 &  0.948$\pm$0.026 &  0.955$\pm$0.032 &  0.976$\pm$0.022 &      1.0$\pm$0.0 &  0.990$\pm$0.004 &  0.998$\pm$0.002 \\
\hline

40  \hspace{0.2cm} Real  &  0.996$\pm$0.005 &  0.993$\pm$0.010 &  0.993$\pm$0.006 &      1.0$\pm$0.0 &      1.0$\pm$0.0 &      1.0$\pm$0.0 &      1.0$\pm$0.0 \\
50  \hspace{0.2cm} Real  &      1.0$\pm$0.0 &  0.992$\pm$0.008 &      1.0$\pm$0.0 &      1.0$\pm$0.0 &      1.0$\pm$0.0 &      1.0$\pm$0.0 &      1.0$\pm$0.0 \\
60  \hspace{0.2cm} Real  &  0.995$\pm$0.007 &  0.993$\pm$0.006 &      1.0$\pm$0.0 &      1.0$\pm$0.0 &      1.0$\pm$0.0 &      1.0$\pm$0.0 &      1.0$\pm$0.0 \\
70 \hspace{0.2cm} Real  &  0.992$\pm$0.008 &      1.0$\pm$0.0 &  0.966$\pm$0.047 &      1.0$\pm$0.0 &      1.0$\pm$0.0 &      1.0$\pm$0.0 &      1.0$\pm$0.0 \\
80  \hspace{0.2cm} Real  &      1.0$\pm$0.0 &  0.996$\pm$0.003 &  0.996$\pm$0.003 &      1.0$\pm$0.0 &      1.0$\pm$0.0 &      1.0$\pm$0.0 &  0.997$\pm$0.002 \\
90 \hspace{0.2cm} Real  &  0.997$\pm$0.002 &      1.0$\pm$0.0 &      1.0$\pm$0.0 &      1.0$\pm$0.0 &      1.0$\pm$0.0 &      1.0$\pm$0.0 &      1.0$\pm$0.0 \\
100  \hspace{0.07cm} Real &  0.995$\pm$0.004 &  0.998$\pm$0.002 &  0.998$\pm$0.002 &      1.0$\pm$0.0 &      1.0$\pm$0.0 &      1.0$\pm$0.0 &  0.998$\pm$0.002 \\
\hline
\end{tabular}
\caption{AUROC reached for synthetic data with various time series lengths and training set sizes}
    \label{f1_synth}
\end{table*}

\subsection{Classification performance over real, regularly sampled time series}
\label{sec:regular}

\begin{table*}[h!]
    \centering
    \caption{AUROC reached by each method over the different experimental scenarios, in case of regular sampling (no missing data).}
    \label{tab:accuracy}
    \small
    \begin{tabular}{c|cccc}
         \textbf{Dataset} &\textbf{ Real data} & \textbf{Time Slicing} & \textbf{Time Warping} & \textbf{T-CGAN}\\
         \hline
         Starlight curves &0.7127 $\pm$ 0.1371 &0.7534 $\pm$ 0.0082 & 0.9840 $\pm$ 0.0099 &\textbf{0.9851} $\pm$ 0.0156 \\
         Power Demand &0.6211 $\pm$ 0.1762 & 0.7152 $\pm$ 0.0932 & 0.7988 $\pm$ 0.0836 & \textbf{0.8336} $\pm$ 0.1553\\
         ECG200 &0.7014 $\pm$ 0.0335& 0.6666 $\pm$ 0.0836 & 0.7227 $\pm$ 0.0391 & \textbf{0.7882} $\pm$ 0.0122 \\
    \end{tabular}
    
\end{table*}

 \begin{table}
 \caption{AUROC values for classification with unbalanced ratio of  $0.1$ for original dataset, and from $0.2$ to $1.0$ for T-CGAN augmented data (Starlight dataset).}
\label{auroc_unb}
 \centering
 \small
\begin{tabular}{c|c|c}
\textbf{Data}&\textbf{Ratio}&\textbf{AUROC}\\
\hline
Original & 0.1 & 0.6798$\pm$0.0222\\
\hline
&0.2 & 0.9574$\pm$0.0082\\
&0.3&  0.9710$\pm$0.0037\\
&0.4&  0.9740$\pm$0.0037 \\
&0.5&  0.9750$\pm$0.0035 \\
Augmented with&0.6&  0.9790$\pm$0.0066 \\
T-CGAN generated&0.7&  0.9780$\pm$0.0059 \\
&0.8&  0.9800$\pm$0.0035 \\
&0.9&  0.9840$\pm$0.0037 \\
&1.0&  0.9851$\pm$0.0024 \\
\hline
\end{tabular}

\end{table}

We now compare T-CGAN against other methods for time series data augmentation, namely time slicing and time warping, over real world datasets. We apply data augmentation in binary classification problems of time series with unbalanced datasets, and we show that our augmentation approach, applied to the class that features less data points, leads to better performance of the classifier.

Notice that in this experiment the sampling interval in the time series is regular. We perform the test also in this setting because  the other methods do not work with irregular time samples. 
The metric chosen for this purpose is the AUROC.
We use the same classification model architecture (CNN) in all cases.

\textbf{Time slicing} is a method inspired by computer vision communities which consists in cutting the time series in slices and performing classification at the slice level. 
During training, the classifier learns how to classify each slice, where the size of the slice is a parameter of the method. At test time, the classification of the time series is performed by classifying each slice taken from the time series and by applying a majority vote over all the slices to decide a predicted label. In our experiment we decide to divide each time series into only 3 slices, because we consider short time series in the first place. In this way, using time slicing, we triple the number of samples.

\textbf{Time warping} consists of warping a randomly selected slice of a time series by adjusting its speed, i.e. by speeding it up or slowing it down. The size of the slice and the warping ratio are parameters of this method. In this paper, we only consider warping ratios equal to $\frac{1}{2}$ or $2$, inspired by the results of \cite{LeGuennec}.

Table \ref{tab:accuracy} shows the performance of the the CNN classifier over each datasets (\ref{real_datasets}): without modification; augmented with time slicing; augmented with time warping; and augmented with T-CGAN. As the table shows, T-CGAN achieves the best classification accuracy in all of the three datasets.

We can observe how the re-balancing influences the classification performance 
in  Table \ref{auroc_unb}. In the table and in the figure we report the AUROC when we re-balance the two classes with the data generated by the GAN in the Starlight dataset.

\begin{table*}[h!]
    \centering
    \caption{AUROC reached by each method over the different experimental scenarios, in case of irregular sampling (20\% missing data, randomly selected), averaged over 10 repetitions.}
    \label{tab:accuracy2}
    \small
    \begin{tabular}{c|cccc}
         \textbf{Dataset} &\textbf{ Real data} & \textbf{Time Slicing} & \textbf{Time Warping} & \textbf{T-CGAN}\\
         \hline
         Starlight Curves &0.6798 $\pm$ 0.0222 &0.5200 $\pm$ 0.0041 & 0.9508 $\pm$ 0.0041 & \textbf{0.9750} $\pm$ 0.0040 \\
         Power Demand &0.5011 $\pm$ 0.0042 &0.5020 $\pm$ 0.1240  &0.5322 $\pm$ 0.0053 & \textbf{0.6999} $\pm$ 0.0356\\
         ECG200 &0.5724 $\pm$ 0.2410 &0.5233 $\pm$ 0.0210  &0.6474 $\pm$ 0.0341& \textbf{0.7202} $\pm$ 0.0546 \\
    \end{tabular}
\end{table*}

\begin{figure*}[h!]
\centering
    \includegraphics[width=.345\textwidth]{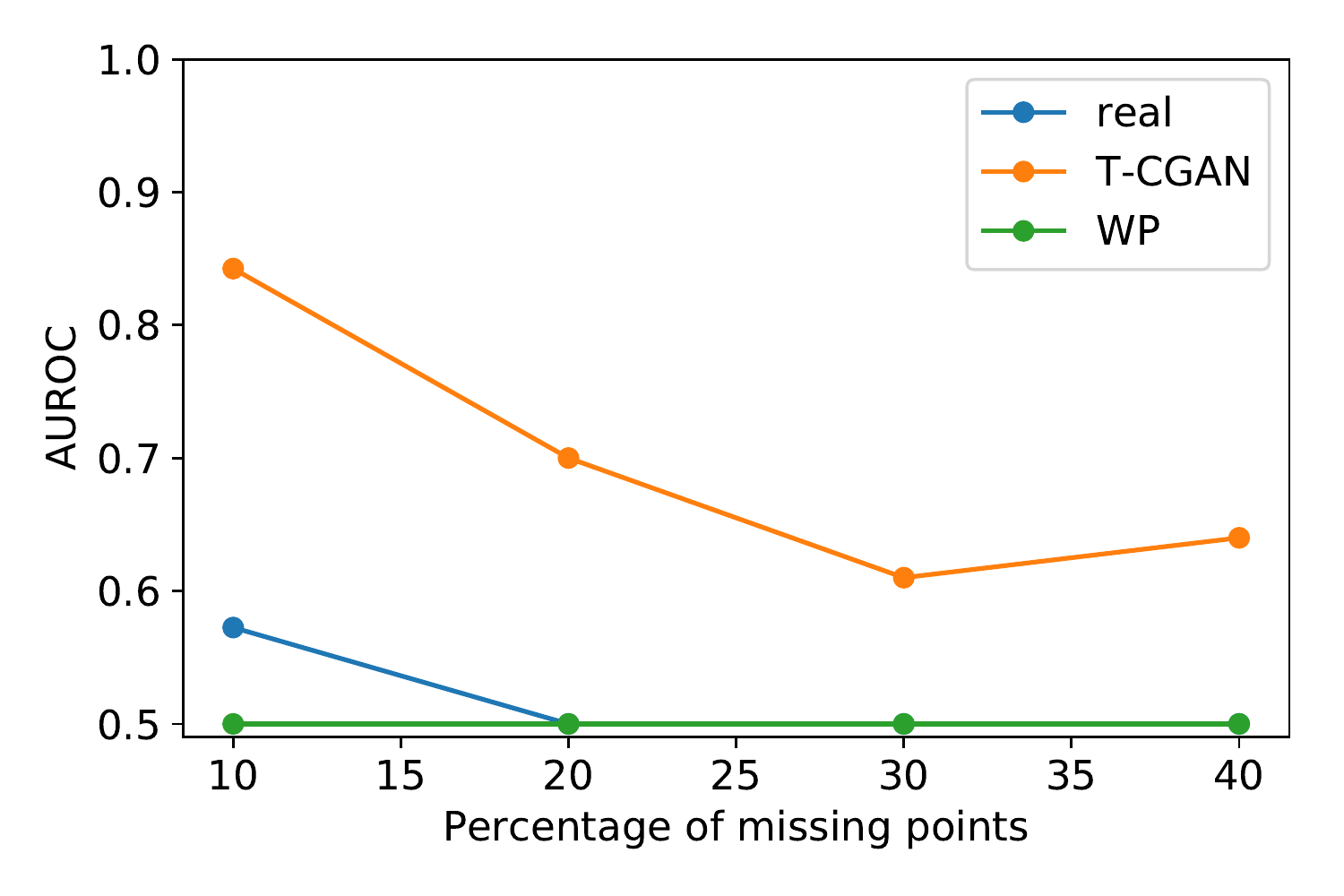} \includegraphics[width=.30\textwidth]{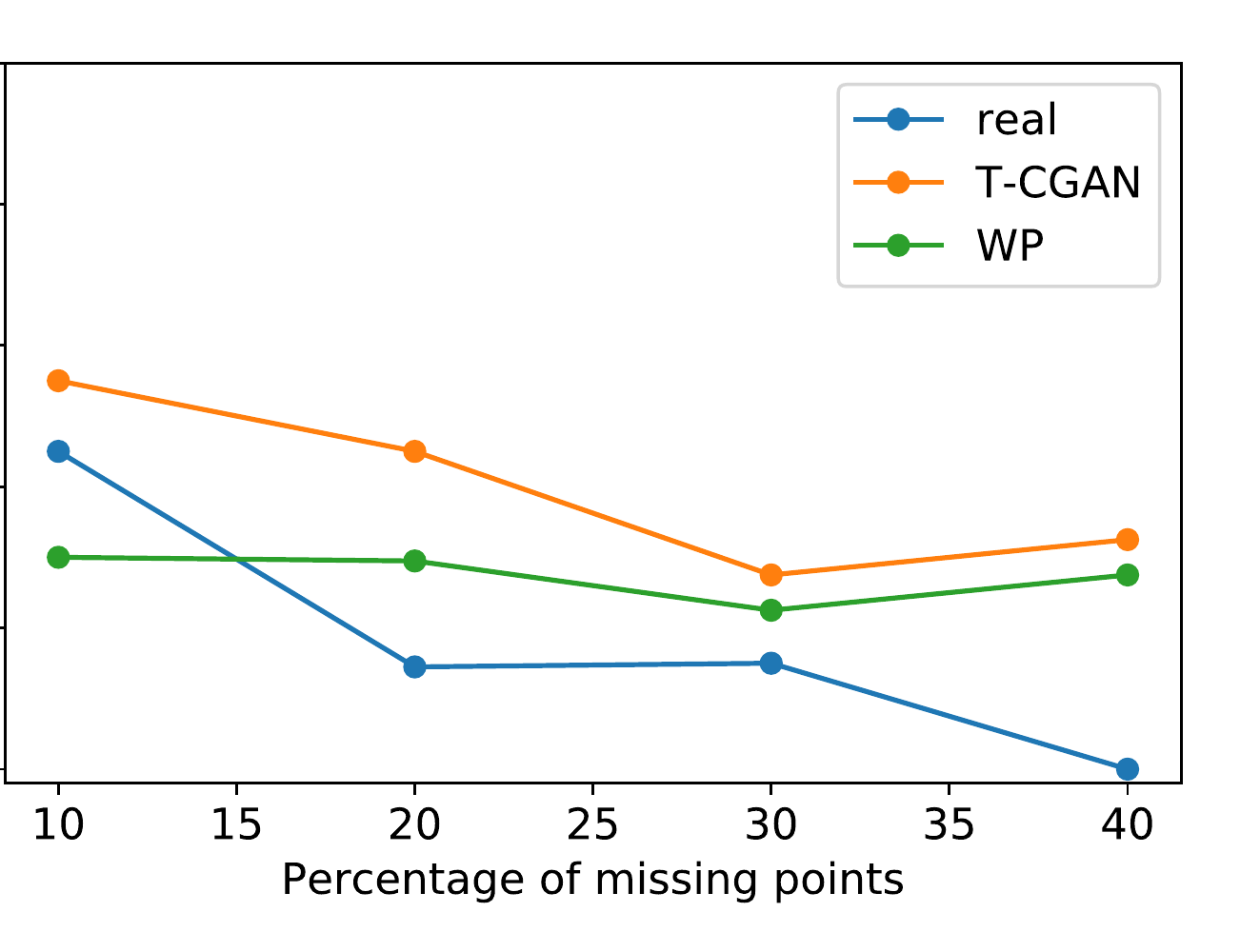} \includegraphics[width=.30\textwidth]{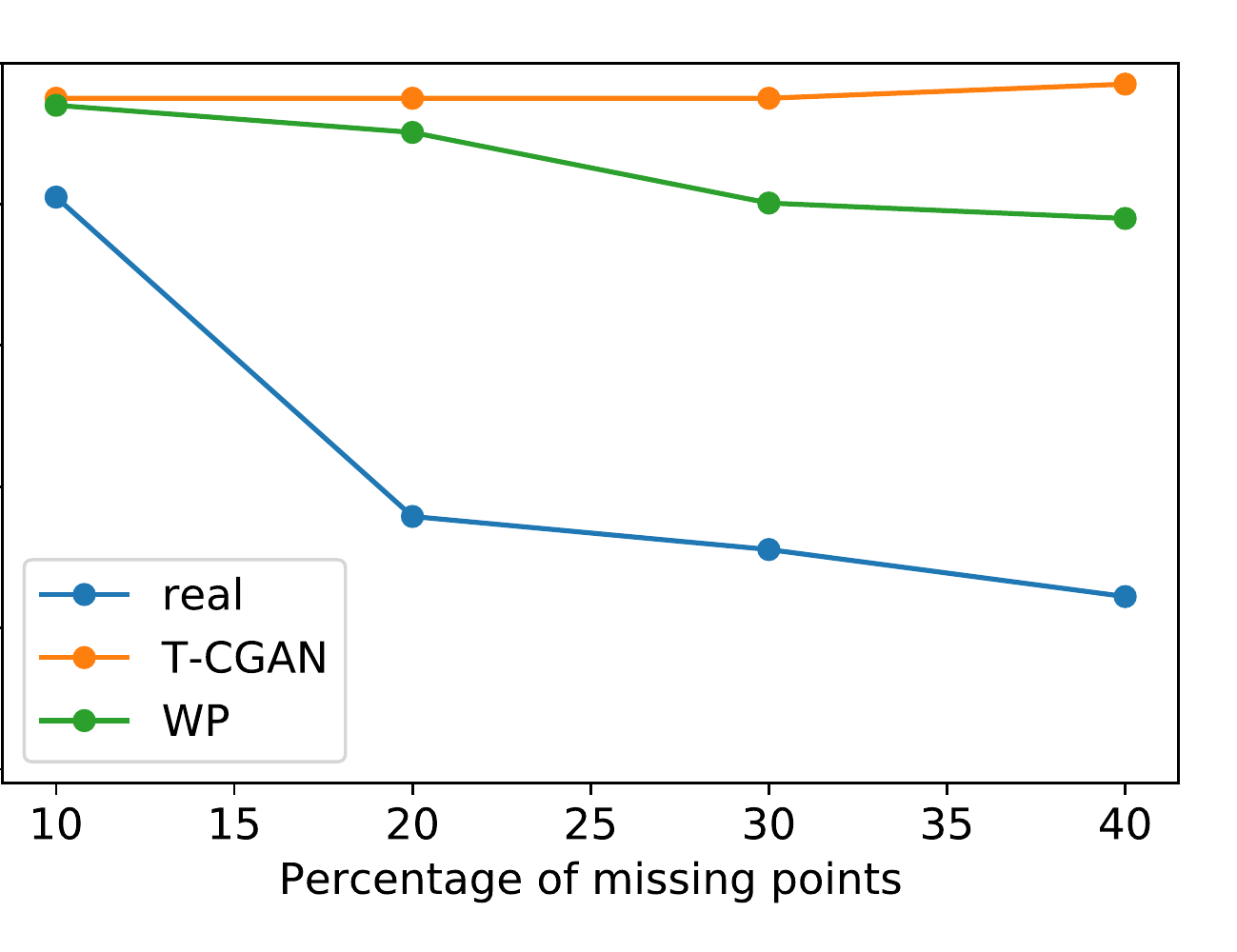}\\
    (a) Power Demand \hspace{65pt}
 (b) ECG200    \hspace{65pt}
 (c) Starlight curves
\caption{
    AUROC with varying percentage (10$\%$, 20$\%$, 30$\%$, 40$\%$)  of missing values for the three datasets without augmentation (real) and with augmentation through time warping  (WP) and T-CGAN (gan).}
    \label{fig:noise_images}
\end{figure*}
\subsection{Classification performance over irregularly-sampled time series}
\label{sec:irregular}

We now analyze the performance of T-CGAN over irregularly-sampled time series. To do so, we generate an irregular version of the real time series by randomly removing 20$\%$ of the data in each curve, thus creating irregular time sampling. 

Again, we compare the performance of the CNN classifier over the datasets (described in Section \ref{real_datasets}): without modification, augmented with time slicing, augmented with time warping, and augmented with T-CGAN. For  time slicing and time warping, before using the classifier, we fill the data for the missing points using interpolation, because these approaches do not support irregularity. 

Table \ref{tab:accuracy2} describes the CNN classifier performance over the three datasets. The table reports a distinct decrease of performance for each of the three methods compared with the performance of Table \ref{tab:accuracy}, as expected due to the missing data points. Also in this setting T-CGAN outperforms the other methods.
T-CGAN maintains good its performance here. Note the performance decrease is dataset-dependent, and varies in particular with the complexity of the time series and its length.

\subsection{Evaluating augmentation over varying percentage of missing values}
\label{sec:imbalance}
Finally, we evaluate our method with respect to different imbalance ratios in the training set, still considering irregularly sampled time series. To this aim, we define an experimental setting with different training sets, built from the original one by removing 10$\%$, 20$\%$, 30$\%$, and 40$\%$ of the training points one of the two classes of the set.  
Figure \ref{fig:noise_images}  shows the classification results for the different dataset with the increasing amount of missing values. 
The figure shows that the impact of the augmentation of the less represented class in the data set is more and more important, when the unbalancing increases. It also shows that T-CGAN outperforms time warping at all levels of unbalancing. 

\section{Conclusion}
\label{sec:concl}
We proposed a generative model for augmentation of time series data with irregular sampling, called T-CGAN. This novel architecture generalizes  conditional GANs so as to deal with the unique characteristics of the irregularly sampled time series. Various experiments with synthetic and real-world datasets show that T-CGAN significantly outperforms state-of-the-art data augmentation techniques on time series, especially for what regards the classification problem, obtaining good accuracy also in case of small training set and short, noisy and irregularly sampled time series.
 In the classification problem over real datasets, our method performs better than any existing approaches to our knowledge 
 in terms of  classification AUROC.  The proposed T-CGAN approach can have significant impact, as it can be helpful in the common scenarios of time series data with irregular sampling, noise, missing points, and insufficient samples.

\bibliographystyle{plain}
\bibliography{biblio.bib}

\end{document}